\pdfoutput=1

\documentclass[11pt]{article}

\usepackage[preprint]{acl}

\usepackage{times}
\usepackage{latexsym}
\usepackage[T1]{fontenc} %
\usepackage[utf8]{inputenc} %
\usepackage{microtype} %
\usepackage{inconsolata} %
\usepackage{graphicx} %

\usepackage{booktabs}
\usepackage{wasysym}

\usepackage{amsmath}
\usepackage{caption}
\usepackage{floatrow}
\usepackage{float}
\usepackage{placeins}
\usepackage{multirow}
\usepackage{makecell}
\usepackage{hyperref}
\usepackage[nameinlink]{cleveref}
\hypersetup{
            pdftitle={Beyond Shallow Heuristics: Leveraging Human Intuition\\ for Curriculum Learning},
            pdfauthor={Vanessa Toborek; Sebastian M\"uller; Tim Selbach; Tam\'as Horv\'ath; Christian Bauckhage},
            pdfkeywords={curriculum learning, simple language, masked language models, llm, nlp},
            bookmarksnumbered=true, 
            pdfcreator={Vanessa Toborek},
            pdfproducer={LaTeX (customized)},
            }
\newcommand{\pvalue}[1]{%
  \ifdim #1 pt < 0.01 pt \textsuperscript{\tiny \eighthnote}%
  \else\ifdim #1 pt < 0.05 pt \textsuperscript{\tiny \eighthnote}%
  \else \textsuperscript{\tiny \,}%
  \fi\fi
}

\newcommand{\bertname}{BERT-tiny}
\newcommand{\baselineEL}{Baseline EL}
\newcommand{\baseline}{Baseline SL+EL}

\title{Beyond Shallow Heuristics: Leveraging Human Intuition\\ for Curriculum Learning}

\author{
 \textbf{Vanessa Toborek\textsuperscript{1,2}},
 \textbf{Sebastian M\"uller\textsuperscript{1,2}},
 \textbf{Tim Selbach\textsuperscript{1}},
 \\
 \textbf{Tam\'as Horv\'ath\textsuperscript{1,2,3}},
 \textbf{Christian Bauckhage\textsuperscript{1,2,3}}
\\
 \textsuperscript{1}University of Bonn,
 \textsuperscript{2}Lamarr Institute,
 \textsuperscript{3}Fraunhofer IAIS
\\
 \vspace{0.2cm}
 \texttt{toborek@cs.uni-bonn.de}
}

\begin{document}
\maketitle
\begin{abstract}
Curriculum learning (CL) aims to improve training by presenting data from ``easy'' to ``hard'', yet defining and measuring linguistic difficulty remains an open challenge. We investigate whether human-curated simple language can serve as an effective signal for CL. Using the article-level labels from the Simple Wikipedia corpus, we compare label-based curricula to competence-based strategies relying on shallow heuristics. Our experiments with a {\bertname} model show that adding simple data alone yields no clear benefit. However, structuring it via a curriculum -- especially when introduced first -- consistently improves perplexity, particularly on simple language. In contrast, competence-based curricula lead to no consistent gains over random ordering, probably because they fail to effectively separate the two classes. Our results suggest that human intuition about linguistic difficulty can guide CL for language model pre-training.
\end{abstract}

\section{Introduction}
\label{sec:introduction}
The growing scale of language models (LMs) has increased interest in training strategies that improve efficiency and convergence. Curriculum learning (CL), inspired by developmental psychology, is one such approach.
CL structures training by presenting examples in a sensible order -- typically from ``easy'' to ``hard''~\cite{Elman1993,Bengio2009,Wang2021survey}.
While intuitively compelling and empirically useful in certain NLP tasks~\cite{Platanios2019,KoichiNagatsuka2021},
its overall impact on masked language model (MLM) pre-training remains debated~\cite{Surkov2022}.

\begin{table}[!htb]
    \centering
    \setlength{\textfloatsep}{1pt}
    \begin{tabular}{p{0.7cm}p{0.7cm}p{5cm}}
    \toprule
    Rarity & Class &  Example\\
    \midrule
    low & SL & She is the author of the Twilight series. \\
    low & EL & The history of poker is the subject of some debate. \\
    \midrule    
    high & SL & Today, most automotive diesels are turbocharged. \\
    high & EL & Pink Floyd watched The Beatles recording Lovely Rita. \\
    \bottomrule
    \end{tabular}
    \caption{Sentences showing examples of high and low average word rarity for each class in the Simple Wikipedia dataset \cite{Kauchak2013}.}
    \label{tab:word-rarity-examples}
\end{table}

A key challenge in CL is the definition of linguistic difficulty. Unlike other domains, language difficulty may arise from multiple dimensions -- such as syntax, semantics or context. In the absence of gold standards, prior work often relies on shallow heuristics~\cite{Platanios2019,Ranaldi2023}. Yet, readability research suggests that no single heuristic reliably captures linguistic complexity~\cite{Battisti2020}. In contrast, humans intuitively consider multiple dimensions when simplifying text. This motivates the central question for this work: \textit{Can human-curated simple language effectively guide CL for MLM pre-training?}

To answer this question, we study CL strategies based on article-level labels from the Simple Wikipedia corpus~\cite{Coster2011} and compare them to \textit{competence-based CL} with shallow difficulty heuristics~\cite{Platanios2019}, using {\bertname} for MLM pre-training.
Our experiments show that merely adding simple language data to training yields no overall improvement. Still, incorporating it through a label-based curriculum consistently improves not only overall perplexity but particularly the simple language perplexity. This effect vanishes when reversed: training on everyday language first is detrimental to learning, underscoring the importance of example ordering. Surprisingly, competence-based curricula show no benefit over random ordering. 

Further, we find that simple and everyday language articles have similar vocabulary sizes and high lexical and distributional overlap on the chosen difficulty heuristics.
This suggests that competence-based CL fails here, because the heuristics do not effectively separate the classes. In contrast, the consistent gains from label-based curricula imply that simple language encodes other useful information, providing structure that benefits pre-training when leveraged correctly.
These results suggest that simple language does indeed help, when applied in a curriculum that makes use of human intuition on linguistic difficulty. 
 
\section{Related Work}
\label{sec:related}
A common form of data-level CL orders the data points according to a global difficulty measure. This approach has been applied to various NLP tasks such as language modelling~\cite{KoichiNagatsuka2021,Ranaldi2023}, machine translation~\cite{Platanios2019,Mohiuddin2022}, and questions answering~\cite{Liu2018} using  difficulty measures like input length~\cite{KoichiNagatsuka2021,Zaremba2015}, word rarity \cite{Platanios2019}, or domain similarity \cite{Mohiuddin2022}. However, the choice of metric is often intuitive and its overall effectiveness remains debated, as the work by~\citet{Surkov2022} found that competence-based CL for MLM offers little to no benefit.

A parallel line of work explores the benefits of simplified language in neural network training. \citet{Mueller2023} show that pre-training on simple language corpora strengthens the syntactic inductive bias in encoder-decoder models.
\citet{Huebner2021} demonstrate that child-directed data facilitates grammar learning for downsized encoder-only models. 
\citet{Lucas2024} explore CL through a masking-based strategy, also leveraging simplified language.
While these studies focus on specific linguistic gains or efficiency improvements, the role of simplified language in global, data-level curriculum design remains unexplored. We address this gap by investigating whether editorially curated simple language -- such as that in Simple Wikipedia -- can serve as an effective learning signal for CL, and how it compares to commonly used difficulty heuristics.

\begin{table}[]
    \centering
    \begin{tabular}{lcc}
    \toprule
    Label & \# tokens & \# sentences \\
    \midrule
    Simple (SL)   & $3,395,297$ & $191,318$ \\
    Everyday (EL) & $3,796,654$ & $176,019$ \\
    \bottomrule
    \end{tabular}
    \caption{Dataset statistics for simple (SL) and everyday (EL) language in the Simple Wikipedia corpus.}
    \label{tab:dataset-stats}
\end{table} 
\section{Methodology}
\label{sec:methodology}
\begin{table*}[h!]
    \centering
\clearpage{}%
\begin{tabular}[t]{lcccc}
    \toprule
    Strategy        & Perplexity             & SL Perplexity          & EL Perplexity           & \# Updates \\
    \midrule
    Baseline EL     & 69.25 \tiny{$\pm$4.04} & 59.50 \tiny{$\pm$4.38} & 81.78 \tiny{$\pm$4.85} & 658\,667 \tiny{$\pm$113\,192} \\
    Baseline SL+EL  & 69.61\pvalue{0.44519} \tiny{$\pm$4.87} & 64.15\pvalue{0.995819} \tiny{$\pm$5.05} & 76.46\pvalue{0.004181} \tiny{$\pm$5.28} & 665\,333 \tiny{$\pm$102\,111} \\
    \midrule
    Incremental     & 66.36\pvalue{0.598140} \tiny{$\pm$2.53}          & 63.29\pvalue{1.000000} \tiny{$\pm$3.39} & \textbf{71.51\pvalue{0.008057} \tiny{$\pm$2.55}} & 781\,333 \tiny{$\pm$83\,312} \\
    Sequential      & \textbf{65.31\pvalue{0.018677} \tiny{$\pm$4.19}} & \textbf{57.83\pvalue{0.000916} \tiny{$\pm$4.52}} & 74.39\pvalue{0.126190} \tiny{$\pm$4.91} & 781\,333 \tiny{$\pm$122\,292} \\    
    Anti-Sequential & 70.32\pvalue{0.252380} \tiny{$\pm$3.97}          & 59.24\pvalue{1.000000} \tiny{$\pm$4.01} & 81.70\pvalue{0.008057} \tiny{$\pm$4.37} & 682\,000 \tiny{$\pm$102\,274} \\
    \midrule
    Length          & 69.05\pvalue{0.890381} \tiny{$\pm$4.15} & 63.84\pvalue{0.977417} \tiny{$\pm$4.12} & 76.37\pvalue{0.899231} \tiny{$\pm$4.46} & 672\,667 \tiny{$\pm$71\,760} \\
    Word Rarity     &  66.74\pvalue{0.890381} \tiny{$\pm$3.48} & 62.48\pvalue{0.977417} \tiny{$\pm$3.52} & 74.12\pvalue{0.718262} \tiny{$\pm$4.12} & 664\,666 \tiny{$\pm$72\,394} \\
    FRE             & 68.05\pvalue{0.778809} \tiny{$\pm$5.22} & 62.53\pvalue{0.977417} \tiny{$\pm$4.98} & 75.32\pvalue{0.899231} \tiny{$\pm$5.88} & 709\,333 \tiny{$\pm$105\,524} \\
    Random          & 68.07\pvalue{0.778809} \tiny{$\pm$4.92} & 63.08\pvalue{0.977417} \tiny{$\pm$4.95} & 75.21\pvalue{0.899231} \tiny{$\pm$5.40} & 679\,333 \tiny{$\pm$105\,388} \\ 
    \bottomrule
\end{tabular}

\clearpage{}%

    \caption{Performance of {\bertname} across baseline and CL strategies. Perplexity is reported for the full dataset and separately for the simple (SL) and everyday language (EL) subsets. Sequential label-based curriculum achieves best overall and SL perplexity. No competence-based strategy shows consistent improvement over baselines. Reported values are mean and standard deviations across 15 runs. {\eighthnote} denotes significant changes.}
    \label{tab:all_results}
\end{table*}

We use the following experimental setup to study the effect of simple language in MLM pre-training.

\paragraph{Dataset}
We employ the Simple Wikipedia dataset~\cite{Coster2011}, the most popular, freely available simple language corpus in English. It consists of articles from the Simple English Wikipedia in simple language (SL) and their counterparts from the English Wikipedia in everyday language (EL). Each sentence inherits the article-level label (SL or EL), which may introduce some label noise due to within-article variation in sentence complexity. \Cref{tab:dataset-stats} compares both classes regarding their respective number of tokens and sentences.

\paragraph{Difficulty Heuristics}
For the competence-based CL, we consider three shallow heuristics for text difficulty: sentence length, word rarity, and the Flesch Reading Ease (FRE) score (cf. ~\citet{Platanios2019}, \citet{Ranaldi2023}). Refer to \Cref{sec:appendix-difficulty} for the details.
In addition to these, we include a random baseline, where difficulty scores are sampled uniformly to isolate the effect of data ordering from the progressive data exposure.

\paragraph{Curriculum Strategies}
We compare two CL paradigms. First, following~\citet{Platanios2019}, we implement the \textit{competence-based} curriculum approach. We sort the training examples according to the aforementioned difficulty measures and gradually expand the training set as model competence increases. The curriculum proceeds until the entire dataset is included. We provide the full implementation details in \Cref{sec:appendix-implementation}. 

Second, we implement two \textit{label-based} curricula using the SL/EL distinction. The sequential strategy first trains on SL until convergence, then continues training on EL. 
To mitigate potential forgetting from fully replacing the training data, we propose an incremental strategy: the model is first trained on SL alone, then continues on the combined SL+EL set, each phase until convergence.
We also include a reverse sequential strategy (first on EL, then SL) as a control strategy.

\paragraph{Training Setup}
We train a {\bertname} model with two transformer layers of hidden size 128, two attention heads, an intermediate feed-forward of size 512, a batch size of eight, and a learning rate of $10^{-4}$. All models are trained until convergence, with early stopping based on validation loss. All experiments are repeated over 15 random seeds to ensure statistical robustness.

\paragraph{Evaluation}
We evaluate model performance using overall perplexity as well as SL and EL subset perplexities. This helps us assess general improvements as well as register-specific gains. Our baselines include models trained with random sampling: one on everyday language only (\baselineEL{}), the other on a uniform mix (\baseline{}). 
\section{Curriculum Learning Results}
\label{sec:results}
We summarise the final performance of the {\bertname} model across all training strategies in \Cref{tab:all_results}, focusing on overall, SL, and EL perplexity, as loss values are less informative.
We compare each strategy against a primary baseline ({\baseline}), trained on SL+EL using random data sampling, with results averaged over 15 seeds. To assess the statistical significance of our results, we apply a one-sided Wilcoxon signed-rank test for symmetric distributions, and a one-sided median bootstrap test otherwise. All $p$-values are adjusted using the Holm-Bonferroni method within each experiment family (baseline, label-based CL, competence-based CL), using $\alpha=0.05$ and directional hypotheses. \Cref{sec:appendix-significance} details the directional hypotheses and the corresponding adjusted $p$-values.

\begin{figure*}[t]
  \BottomFloatBoxes
  \begin{floatrow}
    \ffigbox[0.70\textwidth][][t]
      {\includegraphics[width=\linewidth]{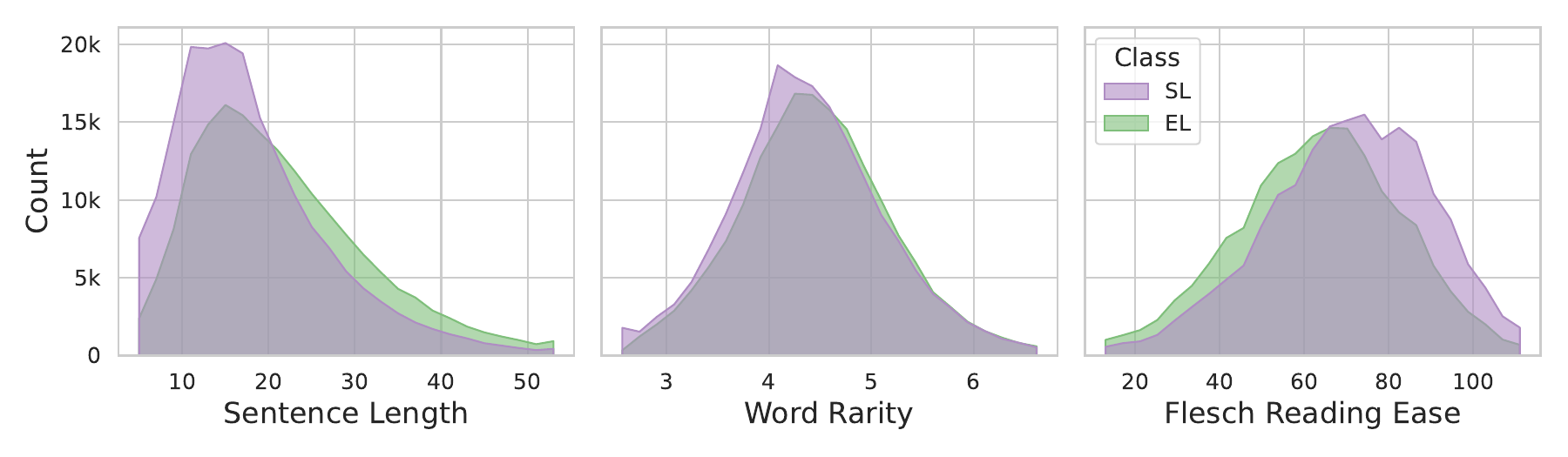}}
      {\caption{Distribution of sentence-level difficulty heuristics for SL and EL. \\ None of the heuristics cleanly separates the two classes.}\label{fig:distributions}}

    \killfloatstyle
    \hspace*{-1.5em}
    \floatbox[{\nocapbeside}]{table}[0.29\textwidth][][t]
      {\begin{tabular}{lrr}
         \toprule
            & SL & EL \\
         \midrule
            SL & $100\%$   & $96.67\%$ \\
            EL & $86.06\%$ & $100\%$   \\
         \bottomrule
       \end{tabular}}
      {\caption{Vocabulary overlap between classes. Over $80\%$ of EL's vocabulary is also present in SL, showing high lexical similarity.}\label{tab:overlap}}
  \end{floatrow}
\end{figure*}

\paragraph{Does merely adding simple language to the training data improve model performance?} The results provide a clear but mixed answer. Comparing {\baseline} to {\baselineEL}, we see a significant improvement in EL perplexity but no improvement in neither overall nor SL perplexity.

\paragraph{Can simple language effectively guide CL?} We find clear evidence in favour of simple language guiding CL -- provided that the sampling strategy is right. Among the label-based CL strategies, only the sequential variant significantly improves overall as well as SL perplexity -- achieving the best scores across all strategies. Incremental improves EL perplexity, but not overall performance. To show that the improvements of the sequential strategy are not accidental, we also test its anti strategy (i.e. starting training on EL, then progressing with SL): it performs similarly to {\baselineEL} and yields significantly worse EL perplexity than {\baseline}. Both incremental and sequential strategies require more updates than {\baseline} to reach these improvements.

\paragraph{Are shallow text features sufficient to guide competence-based CL?} We have a negative answer to this question. Across all three competence-based difficulty measures, we observe no significant improvement in perplexity compared to {\baseline}. The random strategy further suggests that neither simply increasing the dataset size nor imposing an order on shallow features leads to better model performance.

\section{Discussion}
\label{sec:discussion}
In this section we discuss the implications of the results from the previous section with regards to our three research questions.

\paragraph{Learning across registers: asymmetries and interference} The surprisingly strong performance of {\baselineEL} on the SL subset suggests that EL may implicitly cover much of the SL distribution, possibly due to the compositionality of language. However, simply adding SL to the randomly ordered training data does not improve overall performance -- and while it significantly improves EL perplexity, it worsens performance on SL itself. This asymmetry hints at a negative interference effect as observed in multilingual model training \cite{Wang2020a}: though both classes stem from the same language, they might be different enough to cause gradient conflicts when used in the same dataset.
These findings emphasise that learning patterns across language registers are not symmetric, and underscore the importance of evaluating perplexity for different subsets.

\paragraph{Structure matters: the effectiveness of label-based curricula} Models only benefit from SL when introduced in a structured way. Sequential label-based curricula, where training begins with SL before using EL, consistently outperform other strategies in overall and SL perplexity. This aligns with the idea that simplified input can serve as a scaffold, supporting the acquisition of more complex patterns. While the effect mirrors principles observed in human learning, the underlying reason why structured exposure aids generalisation may differ in MLM.
\paragraph{The limits of difficulty heuristics} Competence-based curricula using shallow difficulty heuristics show no clear advantage over random strategies. While this supports prior findings by~\citet{Surkov2022}, our analysis offers further insight. 
\Cref{fig:distributions} shows histograms comparing the distribution of shallow heuristics in SL and EL and \Cref{tab:word-rarity-examples} illustrates some examples.
While it is plausible that EL has samples at the ``easy'' extremes, as not every sentence in everyday language is necessarily complex, we also observe SL examples at the ``complex'' extremes. Assuming that SL represents text that is easier to understand for humans, this highlights that the difficulty heuristics fail to meaningfully separate the two classes.

\paragraph{Future Directions} We find that while shallow difficulty heuristics do not suffice to guide CL, the information encoded in the language classes does. Despite high lexical overlap and comparable size (\Cref{tab:dataset-stats,tab:overlap}), simple language may offer more than surface-level simplicity. Prior work has shown that both humans and neural models benefit from regular, compositional input~\cite{Galke2024a} and simple language might reflect just that through syntactic consistency or clearer discourse structure. Future work could explore how such compositional features manifest in simple language, and whether they can be modelled or annotated as difficulty signals -- enabling broader and more effective CL strategies in MLM pre-training.

\section{Conclusion}
\label{sec:conclusion}
We examined whether human-curated simple language can guide CL in MLM pre-training. Our results show that label-based curricula outperform both random baselines and competence-based approaches relying on shallow difficulty heuristics. While the two language classes show high lexical and distributional overlap, their ordering -- particularly when first training on simple language before moving to everyday language -- leads to significant gains in model performance. 
This suggests that human intuition about linguistic difficulty provides more effective structure for CL than traditional surface-level heuristics.  
\bibliography{custom}

\appendix

\newpage
\section{Implementation Details}
\label{sec:appendix-implementation}

We provide our implementation details for the competence-based CL strategy, where each training sample is assigned a difficulty score and the dataset is sorted accordingly. A predefined competence function then controls the fraction of data available at each training step $t$, gradually increasing the difficulty over time.
Following \citet{Platanios2019}, we adopt the square-root based competence function, which they found to be most effective: $$c_{sqrt}(t)=\min(1, \sqrt{\frac{t(1-c_0^2)}{T}}) \in [0,1],$$ where $c_0$ denotes the initial competence at $t=0$ and $T$ is the total number of steps in the CL phase. In our experiments, we observed that shorter competence phases tend to yield better results than longer ones. We pick $T=50\,000$ and $c_0=0.05$ as function parameters. The size of the training dataset is updated every $5\,000$ steps depending on the current function value.

\section{Difficulty Heuristics}
\label{sec:appendix-difficulty}

In our work we consider three popular heuristics to measure the difficulty of text for global, data-level curriculum learning (cf. \citet{Platanios2019} or \citet{Ranaldi2023}). Let $S$ be a sentence, represented by a finite sequence of words $(w_1, w_2, \dots, w_m)$. The first heuristic, sentence length, is defined by the number of words in the sentence: $$\text{length}(S) = |S|.$$ Next, we use the word rarity metric as proposed by \citet{Platanios2019}, but normalise it by the number of words to remove its strong correlation with the sentence length: $$\text{word\,rarity}(S)=-\frac{1}{|S|}\sum_{w \in S}{\log \left( \frac{\text{count}_c(w)}{N} \right)},$$ where $N$ denotes the size of the vocabulary of the corpus and $\text{count}_c(w)$ the number of times $w$ appeared in the corpus. Last, we present the Flesch Reading Ease (FRE) score as defined by \citet{Flesch1948}. It is designed to evaluate the readability of text and to return a score between 0 and 100: $$\text{FRE}(S)=206.835 - 1.015 \times \text{ASL} - 84.6 \times \text{ASW},$$ where $\text{ASL}$ denotes the average sentence length, which is always the actual sentence length since we only evaluate single sentences, and $\text{ASW}$ denotes the average syllables per word. Since the FRE was designed to evaluate text samples of 100 words, we can encounter negative FRE scores which are outside the originally defined range.

\section{Details on the Significance Tests}
\label{sec:appendix-significance}

\Cref{tab:adjusted-p-values} reports the adjusted $p$-values for all strategies, assessing their performance relative to relevant baselines. For each comparison, we applied a one-sided test based on our directional hypotheses: (1) whether adding SL (\baseline) \textit{improves} over the baseline trained with EL (\baselineEL); (2) whether label-based curricula (Incremental and Sequential) \textit{improve} over the full baseline (\baseline); (3) whether Anti-Sequential \textit{hurts} performance compared to \baseline; and (4) whether competence-based strategies (Length, Word Rarity, FRE, Random) \textit{improve} over the \baseline.

\begin{table}[]
    \centering
\begin{tabular}[t]{p{2cm}ccc}
    \toprule
    Strategy        & PPL             & SL PPL          & EL PPL \\
    \midrule
    Baseline SL+EL  & .445 (w)         & .996 (w)         & \textbf{.004} (w) \\
    \midrule
    Incremental     & .598 (b)          & 1.00 (w)         & \textbf{.008} (w) \\
    Sequential      & \textbf{.019} (w) & \textbf{.001} (w) & .126 (w) \\    
    Anti-Sequential & .252  (b)         & 1.00  (w)        & \textbf{.008} (w) \\
    \midrule
    Length          & .890 (w) & .977 (w) & .899 (w) \\
    Word Rarity     & .890 (b) & .977 (w) & .718 (w) \\
    FRE             & .779 (w) & .977 (w) & .899 (w) \\
    Random          & .779 (w) & .977 (w) & .899 (w) \\
    \bottomrule
\end{tabular}     \caption{Adjusted $p$-values for all statistical tests for the models' performance on overall perplexity (PPL), simple language perplexity (SL PPL), and everyday language perplexity (EL PPL). We choose $\alpha=0.05$ and boldface all significant results. We further indicate which one-sided test was run: (w) Wilcoxon signed-rank test or (b) boostrap median test.}
    \label{tab:adjusted-p-values}
\end{table}

\end{document}